\documentclass{ifacconf}

\usepackage{graphicx}      
\usepackage{natbib}        

\usepackage{multirow} 
\usepackage{booktabs} 
\usepackage{times} 
\usepackage{graphicx} 
\usepackage{dblfloatfix} 

\usepackage{xurl}

\usepackage{amssymb}
\usepackage{amsmath}

\newcommand{\argmin}{\mathop{\mathrm{argmin}}}
\newcommand{\argmax}{\mathop{\mathrm{argmax}}}

\begin{document}
\begin{frontmatter}

\title{Comparing Model-free and Model-based Algorithms for Offline Reinforcement Learning} 

\thanks[footnoteinfo]{The project this paper is based on was supported with funds from the German Federal Ministry of Education and Research under project number 01IS18049A. This work
has been submitted to IFAC for possible publication.}

\author[First]{Phillip Swazinna} 
\author[Second]{Steffen Udluft} 
\author[Third]{Daniel Hein} 
\author[Fourth]{Thomas Runkler}

\address[First]{Siemens Technology / Technical University of Munich \\(e-mail: swazinna@in.tum.de)}
\address[Second]{Siemens Technology (e-mail: steffen.udluft@siemens.com)}
\address[Third]{Siemens Technology (e-mail: daniel.hein@siemens.com)}
\address[Fourth]{Siemens Technology / Technical University of Munich \\(e-mail: thomas.runkler@siemens.com)}


\begin{abstract}
    Offline reinforcement learning (RL) Algorithms are often designed with environments such as MuJoCo in mind, in which the planning horizon is extremely long and no noise exists. We compare model-free, model-based, as well as hybrid offline RL approaches on various industrial benchmark (IB) datasets to test the algorithms in settings closer to real world problems, including complex noise and partially observable states. We find that on the IB, hybrid approaches face severe difficulties and that simpler algorithms, such as rollout based algorithms or model-free algorithms with simpler regularizers perform best on the datasets.
\end{abstract}


\end{frontmatter}

\section{Introduction}
Over the past decade, reinforcement learning (RL) has become a general purpose tool for addressing control problems. It became a well known technique due to success in playing Go better than any human could, playing Atari video games better than most humans, and learning robotic locomotion without any human guidance. While the achievements are remarkable, only rather recently have RL techniques started to address issues more relevant to industrial control problems: Problems such as turbine control, automatic heating / cooling solutions, autonomous driving, and many others often have a distinguishing feature compared to the problems commonly considered in RL literature. Board games, video games, or simulated robots are essentially perfect and deterministic environments, which can also be queried and explored exhaustively without any real cost apart from compute time. \\
In many real-world settings however, usually both of these constraints are violated - practitioners rarely get to run a policy in a simulation, and if yes, it is usually very costly. Interactions with the real environment are even more rare, since they may not only be prohibitively expensive due to decreasing productivity, but may also incur dangerous situations that might brake equipment or even harm workers. Thus, the need for learning control policies directly from previously collected datasets arose. Large datasets also often already exist for many industrial systems since they are usually passively collected during productive operation. E.g. many vehicle manufacturers anonymously collect sensor data during driving for a part of their fleet, and sensor data in industrial control problems such as turbine control is often logged in sub-second intervals, creating huge amounts of data that often goes unused.\\
When traditional (off-policy) RL algorithms are used in these settings, they usually fail due to the narrow data distribution present in the dataset - the value functions or transition models will be evaluated in areas of the state-action space that have rarely or even never been seen in reality, so the assessed performance is inaccurate. During training, policies may seem to be performing well, but when deployed back on the real system, they are actually bad.\\
\textit{Offline reinforcement learning} promises to address this issue of learning from static datasets with possibly narrow data distributions. Usually, algorithms feature a combination of ensembling to make their estimations more conservative, and a regularization of the policy towards the one that generated the dataset, so that models never need to be queried in areas of the state-action space that are unsupported by the dataset. In this work, we compare a representative subset of offline RL algorithms that have been proposed over the past few years. We argue that a key conceptual difference among the methods is whether or not, as well as how they use transition models, which is why we compare model-free, model-based, and hybrid approaches. While most algorithms report performance on MuJoCo domains \citep{todorov2012mujoco}, we argue that these environments do not reflect the characteristics of many real-world tasks since they feature no noise in their transitions. We thus evaluate all algorithms on the industrial benchmark (IB) \citep{hein2017benchmark}, an environment motivated by industrial control problems that features complex, multimodal and heteroscedastic noise, to better assess their suitability in practice.

\section{Related Work}

Recently, offline RL has taken off as a topic and numerous papers about a variety of the problem's aspects have since been published. Among the \textbf{early works} in offline RL are a number of model-free algorithms that already considered learning purely from datasets, however had unrealistic assumptions about how those were collected - usually with uniformly random actions. Among those works are Least Squares Policy Iteration (LSPI) \citep{lagoudakis2003least}, Fitted Q Iteration (FQI) \citep{ernst2005treebased}, Neural Fitted Q iteration (NFQ) \citep{riedmiller2005neural}, as well as Neural Rewards Regression (NRR) \citep{schneegass2007improving}. The methods are sometimes also referred to as being semi-batch (or semi-offline) methods, since they sometimes do collect additional data from the real system, however then go back and use all of the so far collected data as a sort of growing batch for learning. \cite{RCNN2007} was among the first model-based offline RL methods, however the considered datasets were still collected largely at random, i.e. with very good exploration. Other, more recent approaches may also address the offline setting, however their focus is often something else: \citep{hein2016reinforcement,hein2018interpretable} focus on finding interpretable policies that are able to increase trust brought towards them by practitioners, while \citep{depeweg2016learning,depeweg2017decomposition,kaiser2020bayesian} put their emphasis on modeling the complicated uncertainties in the transition dynamics of the environments. While theoretically being offline, these algorithms also assume randomly collected datasets.\\

\textbf{Model-free} methods constitute the majority of offline RL algorithms in current RL literature. Among the first algorithms that considered the problem now known as offline RL - with no environment interaction and learning only from a static dataset that was collected under a baseline policy - was Safe Policy Improvement under Baseline Bootstrapping (SPIBB) \citep{laroche2017safe}. It was designed for discrete actions and assumes that the baseline policy is passed to the learning algorithm as an input. Batch Constrained Q-learning (BCQ) \citep{fujimoto2018off} was then among the first offline algorithms for environments with continuous state and action spaces. It also does not rely on the behavior policy being known to the learning algorithm, but instead learns a representation of it in form of a variational autoencoder, which can in turn be used for sampling from the behavior policy. Even though BCQ learns a closed-form perturbation model, it can still be seen as a form of Q-learning since it uses the maximum operator over sampled actions in the Bellman Q-value backup. Expected Max Q-learning (EMaQ) \citep{ghasemipour2021emaq} simplifies the BCQ methodology and leaves out the policy altogether. Bootstrapping Error Accumulation Reduction (BEAR) \citep{kumar2019stabilizing} was among the first actor-critic methods in the offline RL setting with narrow data distributions. It learns a closed form policy, by penalizing the maximum mean discrepancy between actions performed by the learned, and actions sampled from the behavior policy. Behavior Regularized Actor Critic (BRAC) \citep{wu2019behavior} is a generalization of BCQ and BEAR, and introduces the idea of not only penalizing the policy improvement step, but also the learning of the value function. BRAC thus penalizes KL-divergences between learned and behavior policy in both steps. The Advantage weighted Behavior Model (ABM) \citep{siegel2020keep} follows a very simple motto: Keep doing what worked. The algorithm learns a prior for the new policy, where trajectory parts that yielded higher returns are more likely to be reproduced. AlgaeDICE \citep{nachum2019algaedice}, GenDICE \citep{zhang2020gendice}, and GradientDICE \citep{zhang2020gradientdice} build on the DICE (DIstribution Correction Estimation) framework introduced in \citep{nachum2019dualdice} and estimate a ratio that corrects for the discrepancy between the stationary and empirical distributions of the learned policy and the initial dataset. In \citep{agarwal2020optimistic}, the authors introduce the Random Ensemble Mixtures (REM) algorithm, that regularizes learning by requiring the policy to be robust and perform well on any of the learned ensemble members that represent a value function. Similarly, PEBL (PEssimistic ensemBLes for offline deep reinforcement learning) \citep{smit2021pebl} build on Double Deep Q-learning (DDQ) and Soft Actor-Critic (SAC) \citep{van2016deep,haarnoja2018soft} and estimate uncertainty using a multi-headed bootstrap approach to calculate an effective pessimistic value penalty. Policy Sampling Error Corrected TD-0 (PSEC-TD-0) \citep{pavse2020reducing} addresses the issue that in the usual temporal difference estimation of value functions from batch datasets, the updates are weighted by how often an action occurred in the dataset, rather than how often it would occur under the trained policy. The algorithm uses importance sampling to correct the mismatch. \citep{liu2020provably} introduce the family of Marginalized behavior supported algorithms. They find that the usual assumptions on concentrability in the datasets are too strong for practical applications, and modify the Bellman backup to be more conservative to find the approximately best policy within the space explored by the behavior policy. Similarly, Conservative Q-Learning (CQL) \citep{kumar2020conservative} aims to learn a conservative Q-function so that its values lower bound the true performance of the corresponding policy. Best Action Imitation Learning (BAIL) \citep{chen2019bail} is one of the very few algorithms training a state-value function to find good actions to imitate with a policy network. In Critic Regularized Regression (CRR) \citep{wang2020critic}, taking actions outside the training distribution is discouraged by the use of filtered policy gradients: The new policy is trained to imitate behavior seen in the initial dataset, however the behavior is filtered by the Q-function, so that promising actions are more likely to be imitated than others. Similarly, in Curriculum Offline Imitation Learning (COIL) \citep{liu2021curriculum}, the current policy is improved by an experience picking strategy to imitate from adaptive neighboring policies with a higher return, leveraging that imitation learning can imitate neighboring policies of the behavioral with fewer samples than usual. In Offline Risk Averse Actor Critc (O-RAAC) \citep{urpi2021risk}, a risk averse criterion is learned, which gives the algorithm theoretical performance guarantees with respect to certain distributional shifts. The authors show that their approach leads to much fewer high risk state visitations than other offline algorithms that simply optimize for average performance. In OPAL (Offline Primitive discovery for Accelerating offline reinforcement Learning) \citep{ajay2020opal} the authors address problems with long horizons and sparse rewards. By imitating trajectory parts that are distilled to primitives, and then learning a meta policy to choose among the primitives, the algorithm is able to effectively shorten the actual horizon and learn much faster in the offline sparse reward setting. \citep{shrestha2020deepaveragers} introduce the concept of Deep Averagers with Costs MDPs (DAC-MDPs), which can account for limited data by introducing costs for exploiting under-represented parts of the model, and theoretically lower bound the policy's performance. \citep{fujimoto2021minimalist} introduces TD3+BC (Twin Delayed Deep Deterministic policy gradient + Behavior Cloning), a minimalist algorithm that does not even contain a model of the behavioral policy. Inspite of its simplicity, it surprisingly achieves state of the art performance on benchmark tasks.\\

\textbf{Model-based} offline methods learn a transition model based on the environment interactions observed in the dataset. Since transition models are attributed to increase the data efficiency of RL algorithms, it seems straightforward that they could be useful in the naturally data scarce offline setting. A distinguishing factor among model-based methods is how the transition models are then used for policy training, and whether or not another value function is being used in the process. We found mainly two families of methods, both generally for offline, but also for online RL in the context of model-based approaches: The first way is to use the learned transition model directly and exclusively for policy training. 
In MOdel-based Offline policy Search with Ensembles (MOOSE) \citep{swazinna2021overcoming} for example, the current policy candidate is rolled out through the transition model, and the policy is then improved by calculating the gradient of the resulting return with respect to the policy parameters, as introduced in \citep{RCNN2007}. Similarly, in weight space behavior constraining (WSBC) \citep{swazinna2021behavior}, the policy is searched with a gradient-free, population-based algorithm, and the policy candidate's fitness is determined by performing virtual rollouts through the model. On the other hand, methods like MOReL (Model-based Offline Reinforcement Learning), MOPO (Model-based Offline Policy Optimization), and COMBO (Conservative Offline Model-Based policy Optimization) \citep{kidambi2020morel,yu2020mopo,yu2021combo} use the transition models exclusively to produce additional data samples to augment the initial dataset. The actual policy learning is in those cases still dependent on a value function, which is why we will refer to these methods in this paper as being \textbf{hybrids}. MOReL and MOPO both employ Gaussian transition models and leverage the resulting uncertainty estimates to regularize either the sampling or the rewards, so that the policy does not exploit the models in regions of the state-action space in which they are inaccurate. COMBO is able to regularize the policy without explicitly considering model uncertainty - instead, it simply uses the distance (in steps) from real to synthetic data samples as a heuristic.

\section{Prior Assumptions / Problem Definition}
Formally, we would like to optimize the long term reward in a Markov decision process (MDP) $\mathcal{M}=<\mathcal{S}, \mathcal{A}, \mathcal{T}, \mathcal{R}, \mathcal{s}_0, \gamma, H>$, where $\mathcal{S}$ is the set of states, $\mathcal{A}$ is the set of actions, $\mathcal{T}:\mathcal{S}\times\mathcal{A}\rightarrow\mathcal{S}$ are the transition dynamics, $\mathcal{R}:\mathcal{S}\times\mathcal{A}\rightarrow\mathcal{R}$ is the reward function, $\mathcal{s}_0$ is the set of starting states, $\gamma$ is the discount factor and $H$ the horizon (aka trajectory length). We seek to optimize the Return $R = \sum_{t=0}^H \gamma^t r_t$ by learning a policy $\pi(\cdot)$ that maps from states to actions (either stochastically or deterministically).\\
In offline RL as opposed to the normal, online setting, we are however not allowed to interact with the MDP $\mathcal{M}$. Instead, we are provided with a dataset of $N$ transitions $\mathcal{D}=\{s_t, a_t, r_t, s_{t+1} \}_{t=0}^{N}$, that was created while running a behavior policy $\beta(\cdot)$. We are not provided with the actual $\beta(\cdot)$ and need to learn the policy $\pi(\cdot)$ purely from the dataset $\mathcal{D}$ and may deploy only one final policy.

\section{Considered Methods}
As it is the case for all reinforcement learning algorithms, a natural division among algorithms in offline RL is whether or not the method uses a transition model during training of the policy. In this work we will also examine how the transition model is used, and consider hybrid approaches that employ both value functions as well as transition models.
\subsection{Model-free}
Model-free methods abstain from training a transition model from the action and observation data, and instead aim to directly assign a value to either a state or a state-action pair. The in the following considered offline RL algorithms all train an action-value function aka Q-function, or an ensemble of them. While Q-learning approaches would optimize the Q-function using the Bellman optimality 
operator
\begin{equation}
    \mathcal{B}^* Q(s,a) = r(s,a) + \gamma \mathrm{E}_{s' \sim P(s'|s,a)} [\max_{a'} Q(s',a')]
\end{equation}
the considered approaches are all actor-critic methods, which alternate between estimation of the current policies Q-function and a policy improvement step. They thus employ the regular Bellman operator
\begin{equation}
    \mathcal{B}^{\pi} Q(s,a) = r(s,a) + \gamma \mathrm{E}_{s' \sim P(s'|s,a); a' \sim \pi} [Q(s',a')].
\end{equation}
In model-free Offline RL, we neither know the true transition probabilities P, nor do we want to estimate them from data. Corresponding algorithms thus directly operate on the samples collected in the provided dataset $\mathcal{D}$, and optimize the Q-function by applying the Bellman operator and minimizing the single step temporal difference error:
\begin{align}
    \label{basic_Q}
    \hat{Q}^{k+1} = & \argmin_Q \mathrm{E}_{s, a, s' \sim \mathcal{D}; a' \sim \pi} [ r(s,a)\\
    & + \gamma \mathrm{E}_{a' \sim \pi(a'|s')}[\hat{Q}^k(s', a')] - Q(s,a) ]^2 \nonumber
\end{align}
All algorithms use an ensemble of Q-functions to calculate the Q-target $\hat{Q}^k(s',a')$ by various weighting schemes of the ensemble members, and consequently also find multiple resulting $\hat{Q}^{k+1}$. The usual update rule for policy improvement in actor-critic methods would then be:
\begin{equation}
    \pi^{k+1} = \argmax_{\pi} \mathrm{E}_{s \sim \mathcal{D}; a \sim \pi(a|s)}[\hat{Q}^{k+1}(s,a)]
\end{equation}
The presented algorithms however all adjust this update rule, some of them even alter the Q-function update, all with the target of regularizing policy candidates towards the behavioral policy $\beta(\cdot)$ which generated the dataset $\mathcal{D}$. In the following, we highlight the key conceptual changes the Offline algorithms make to this standard approach - implementational details may be left out due to clarity and brevity.\\

\textbf{Batch Constrained Q-learning (BCQ)} could be considered a sort of hybrid method between Q-learning and actor-critic paradigms: It employs a variational autoencoder (VAE) $\omega(\cdot)$ to model the behavioral policy $\beta(\cdot)$, which is in turn used to sample likely actions to maximize over during the Q-function update, making it seem more like a Q-learning method. On the other hand, BCQ also trains a policy of some sort, which the authors call perturbation model $\xi$, that is allowed to alter the sampled action by a small amount $\Phi$. Since the sampling together with the perturbation model can be viewed as a hierarchical policy, BCQ could also be considered an actor-critic method:
\begin{align}
    \hat{Q}^{k+1} = & \argmin_Q \mathrm{E}_{s, a, s' \sim \mathcal{D}} [ r(s,a)\\
    & + \gamma \max_{a' \sim \omega(s')}[\hat{Q}^k(s', \xi^k(s',a'))] - Q(s,a) ]^2 \nonumber \\
    \xi^{k+1} = & \argmax_{\xi} \mathrm{E}_{s \sim \mathcal{D}; a \sim \omega(s)} [\hat{Q}^{k+1}(s, \xi(s,a))]
\end{align}

\textbf{Bootstrapping Error Accumulation Reduction (BEAR)} also uses a variational autoencoder $\omega(\cdot)$ as an explicit model of the behavior policy. However, BEAR can more clearly be categorized as being actor-critic, since $\omega(\cdot)$ is not used for sampling and maximizing over during the Q-function update, as BEAR trains a closed form policy. While the Q-function update thus looks like equation \ref{basic_Q}, the policy update rule needs to be adjusted since otherwise the policy will want to perform actions for which the Q-function cannot provide an accurate value estimate as it is lacking support in the dataset. BEAR thus penalizes the policy for large maximum mean discrepancies (MMD) between trained policy $\pi(\cdot)$ and learned behavior policy $\omega(\cdot)$:
\begin{align}
    \pi^{k+1} =& \argmax_{\pi} \mathrm{E}_{s \sim \mathcal{D}; a \sim \pi} [\hat{Q}^{k+1}(s,a) \\
    & - \lambda \cdot MMD(\pi(\cdot|s), \omega(\cdot|s))] \nonumber
\end{align}

The \textbf{Behavior Regularized Actor Critic (BRAC)} framework brings two families of algorithms to the table. BRAC-p regularizes only the policy update via a KL-divergence based penalty, while BRAC-v extends the penalty to also regularize the trained state-action value functions. The resulting Q- and policy updates are given by:
\begin{align}
    \hat{Q}^{k+1} = & \argmin_Q \mathrm{E}_{s, a, s' \sim \mathcal{D};a' \sim \pi(s')} [ r(s,a) - Q(s,a)\\
    & + \gamma[\hat{Q}^k(s', a') - \alpha \rm{KL}(\beta(\cdot|s')||\pi(\cdot|s'))] ]^2 \nonumber \\
    \pi^{k+1} = & \argmax_{\pi} \mathrm{E}_{s \sim \mathcal{D}} [\hat{Q}^{k+1}(s,a) - \alpha \rm{KL}(\beta(\cdot|s)||\pi(\cdot|s))]
\end{align}
BRAC uses a simple Gaussian feedforward policy trained via negative log likelihood to model the generating policy $\beta(\cdot)$. BRAC-p is obtained by setting $\alpha=0$ in the Q-function update, we will however only consider BRAC-v in our experiments, since the authors find that version to perform best.

\textbf{Conservative Q-Learning (CQL)} revolves around the idea of learning a conservative estimate $\hat{Q}^{\pi}$ of the value function, which lower bounds the true value of $Q^{\pi}$ at any point. The authors show that this can effectively circumvent the overestimation bias in unexplored regions of the state space that is the most common issue in offline RL. The algorithm aims to achieve this by optimizing the value function to not just minimize the temporal difference error based on the interactions seen on the dataset, but also by minimizing the value of actions that the currently trained policy takes, while at the same time maximizing the value of actions that the behavioral policy took during data generation:
\begin{align}
    \hat{Q}^{k+1} =& \argmin_Q \alpha \cdot (\mathrm{E}_{s \sim \mathcal{D}; a \sim \pi(a|s)}[Q(s,a)] - \mathrm{E}_{s,a \sim \mathcal{D}}[Q(s,a)]) \nonumber \\
    &- \frac{1}{2} \mathrm{E}_{s,a,s' \sim \mathcal{D}}[(Q(s,a) - \mathcal{B}^{\pi}\hat{Q}^k)^2]
\end{align}

\textbf{Twin Delayed Deep Deterministic policy gradient + Behavior Cloning (TD3+BC)} is the most recent and state of the art model-free algorithm considered in our experiments. In contrast to the previous three methods, it does not explicitly learn a model of the behavioral policy. Instead, it directly penalizes Euclidean distance to the actions that were recorded in the dataset. The Q-update is thus equivalent to equation \ref{basic_Q}. The policy is updated using:
\begin{equation}
    \pi^{k+1} = \argmax_{\pi} \mathrm{E}_{s,a \sim \mathcal{D}} [\lambda\hat{Q}^{k+1}(s,\pi(s)) - (\pi(s) - a)^2]
\end{equation}

\subsection{Model-based}
Model-based reinforcement learning methods explicitly train a transition model $P(s'|s,a)$, which can then be used to sample synthetic new data points. Since model-based methods are generally said to be more sample efficient, the argument exists that they consequently need to be better suited for offline RL setting, as we expect to have a naturally data-scarce situation. Various algorithms thus try to leverage the advantage of the transition models to perform better in the offline setting.

\textbf{MOdel-based Offline policy Search with Ensembles (MOOSE)} trains an ensemble of deterministic feedforward transition models $f^n$ via mean squared error of the normalized state deltas as well as a model of the behavioral policy by means of a variational autoencoder $\omega(\cdot)$ before the actual policy training begins:
\begin{align}
    f^n = &\argmin_f \mathrm{E}_{s,a,s' \sim \mathcal{D}} [f(s, a) - \frac{(s' - s) - \mathbf{\mu^{\Delta s}}}{\mathbf{\sigma^{\Delta s}}}]^2 \\
    \omega = &\argmin_{\omega} \mathrm{E}_{q_{\omega}(z|s,a)} [- \rm{log} p_{\omega}(s,a|z) \nonumber \\
    &+ \rm{KL}(q_\omega(z|s,a)||p(z))]
\end{align}
The policy is then optimized by performing gradient descent on rollouts of a fixed length $H$ through the transition models. At the same time, the policy needs to be regularized to be close to the behavioral one, so the likelihood under the autoencoder model (i.e. its reconstruction error), is used as a penalty term:
\begin{align}
    \pi^{k+1} = \argmax_{\pi} \mathrm{E}_{s_0 \sim \mathcal{D}; a_t \sim \pi(s_t); s_t \sim f(s_{t-1}, a_{t-1})} \\
    \left[\sum_{t=0}^H \gamma^t r(s_t, a_t) - (a_t - \omega(s_t))^2 \right] \nonumber
\end{align}

The \textbf{Weight Space Behavior Constraining (WSBC)} algorithm is conceptually similar to MOOSE, however it does not regularize the newly trained policy to be close to the behavioral one in the action space - instead the algorithm constrains trained policies implicitly to be similar to the behavioral by means of a constraint in the policy weight space. The rollouts are also conducted with deterministic transition models, however in WSBC they are recurrent instead of simple feedforward networks. This leads to:
\begin{align}
    \pi^{k+1} = \argmax_{\pi} \mathrm{E}_{s_0 \sim \mathcal{D}; a_t \sim \pi(s_t); s_t \sim f(h_{t-1}, a_{t-1})} \\
    \left[\sum_{t=0}^H \gamma^t r(s_t, a_t) \right] s.t. \quad ||\theta^{k+1} - \phi|| < \epsilon \nonumber \\
    \rm{where} \quad \beta_{\phi} = \argmin_{\beta} \mathrm{E}_{s,a \sim \mathcal{D}} [\beta(s) - a]^2
\end{align}

\subsection{Hybrid Methods}
Commonly, algorithms are labeled model-based as soon as they contain a transition model somewhere. We will however make an effort to further distinguish between model-based methods: The previous section featured methods, that directly used their transition models for policy search, without any extra steps and without the data produced during virtual rollouts being used again. Contrary to this paradigm, there exist approaches that use the transition model only in between policy improvement steps to sample more transitions. These methods thus actually feature a model-free, value function based algorithm with extra data generated by a transition model. These methods try to unify the best of both worlds: Low bias from the model-free, and high data efficiency from the model-based domain. We will consequently refer to these approaches as \textit{hybrids}.

\textbf{Model-based Offline Reinforcement Learning (MOReL)} trains feedforward transition models similar to those employed in MOOSE, however they are not directly used for rollouts and policy training. Instead, the policy is trained via a model-free, Q-function based algorithm (NPG or TRPO) \cite{kakade2001natural, schulman2015trust} on the available data samples. After a certain number of policy update steps, the policy is used to obtain new synthetic data samples by rolling out the policy through the transition models. The newly acquired data is then used to augment the original dataset, and the model-free policy training starts again. The regularizing factor in this algorithm is called unknown state detector (USD) - it prevents the rollouts from entering areas in the state space that are too unknown to the transition models:
\begin{equation}
    \rm{USD}(s,a) = \left( \max_{i,j} [f_i(s,a) - f_j(s, a)]^2 \right) < \epsilon
\end{equation}

\textbf{Model-based Offline Policy Optimization (MOPO)} trains an ensemble of stochastic transition models with Gaussian outputs from the initial datasets. The ensemble members are then used to augment collected rewards, by subtracting a penalty term based on the estimated uncertainty:
\begin{equation}
    \Tilde{r}(s,a) = \hat{r}(s,a) - \lambda \max_{i=1..N} ||\Sigma_{\phi}^i (s,a)||_F
\end{equation}
Where $\hat{r}(s,a)$ is the mean of the rewards predicted by the ensemble, and $\Sigma_{\phi}^i(s,a)$ is the predicted standard deviation of ensemble member $i$. The algorithm thus penalizes the maximum predicted uncertainty by any of the ensemble members. \\
MOPO repeatedly performs short rollouts through the transition models and adds the collected interactions to its replay buffer. The replay buffer can then be used by any off-policy algorithm for policy improvement steps, until the policy is again deployed for further data collection. In their own implementation, the authors use SAC, a model-free, value-based algorithm, to optimize over the models.

\section{Experiments}
\subsection{Datasets}
We evaluate the presented algorithms on the industrial benchmark datasets that were initially proposed in MOOSE. The sixteen datasets feature three different baseline policies mixed with varying degrees of exploration. Together, they constitute a diverse set of example settings in which Offline RL algorithms have to prove themselves in practice: The optimized baseline was obtained by model-based RL algorithm GPRL and simulates the expert case, where practitioners have already obtained a close to optimal controller. The mediocre baseline simply tries to steer the system to a fixed point that is rather well behaved, while the bad baseline steers to a point in which rewards are extremely bad. The exact policies are given by formula \ref{baselines}. The amount of uniform random actions $\varepsilon$ varies from 0\% to 100\%, making the $\varepsilon=0.0$ datasets also extreme cases of the narrow distribution problem.

Note that the input variables $(\Tilde{p}, \Tilde{v}, \Tilde{g}, \Tilde{h}, \Tilde{f} , \Tilde{c})$ are normalized by subtracting their respective mean $(55.0, 48.75, 50.53, 49.45,$ $37.51, 166.33)$ and dividing by their respective standard deviation $(28.72, 12.31, 29.91, 29.22, 31.17, 139.44)$.

\begin{small}
\begin{align}
\label{baselines}
    \pi_{\rm{bad}} &= \begin{cases} 100 - v_t\\ 100 - g_t\\ 100 - h_t \end{cases},
    \pi_{\rm{med}} = \begin{cases} 25 - v_t\\ 25 - g_t\\ 25 - h_t \end{cases}, \\
    \pi_{\rm{opt}} &= \begin{cases} \hfil -\Tilde{v}_{t-5} - 0.91\\ \hfil 2 \Tilde{f}_{t-3} - \Tilde{p} + 1.43\\ 
    -3.48 \Tilde{h}_{t-3} - \Tilde{h}_{t-4} + 2 \Tilde{p} + 0.81 \end{cases} \nonumber
\end{align}
\end{small}

\textit{We make the datasets publicly available for future work under} \url{https://github.com/siemens/industrialbenchmark/tree/offline_datasets/datasets}

\subsection{Evaluation}
During policy training with the offline RL algorithms, we regularly evaluate the current policy candidate on the true benchmark, and calculate a performance by averaging over ten rollouts. Out of these values, we select for each run of an algorithm on a dataset the last 10\%. We combine the values from five seeds and calculate the tenth percentile performance as well as its standard error and report them in Table \ref{table:performance}. This form of evaluation is closer to a worst-case analysis than the typical average performance reporting. In this case, the conservative estimation is important, since there exists no equivalent to early stopping from supervised learning in the offline RL setting: Offline policy evaluation and selection are still open problems \citep{hans2011agent, paine2020hyperparameter, konyushkova2021active, zhang2021autoregressive, fu2021benchmarks}, so we need to stop the policy training at some random point and deploy that policy on the real system. The above procedure is meant to simulate this random stopping, and should quantify how good the algorithm performs at least. We argue that reporting mean results would be much less appropriate, since it doesn't reflect the risk that a practitioner takes if a selected policy performs much worse than average.

\setlength{\tabcolsep}{1.15pt}
\begin{table*}
\caption{\label{table:performance} Tenth percentile performances of evaluated Offline RL algorithms and their standard error. Results for algorithms marked with $\dagger$ are taken from \citep{swazinna2021overcoming}, and results marked with $\ddagger$ from \citep{swazinna2021behavior}.}
  \begin{center}
    \scriptsize
    \makebox[0pt]{\begin{tabular}{|ccc|ccc||ccrlcc|ccrlcc|ccrlcc|ccrlcc|ccrlcc|ccrlcc|} 
    \toprule
       \multicolumn{6}{|c||}{$\varepsilon = $} & \multicolumn{6}{c|}{0.0} & \multicolumn{6}{c|}{0.2} & \multicolumn{6}{c|}{0.4} & \multicolumn{6}{c|}{0.6} & \multicolumn{6}{c|}{0.8} & \multicolumn{6}{c|}{1.0} \\
      \midrule
      & \parbox[t]{2mm}{\multirow{9}{*}{\rotatebox[origin=c]{90}{Bad}}} &
     & & BRAC-v$^\dagger$ & & & & -274 & (12) & & & & & -270 & (12) & & & & & -199 & (7) & & & & & -188 & (8) & & & & & -140 & (5) & & & & & & & & \\
     & & & & BEAR$^\dagger$ & & & & -322 & (4) & & & & & -168 & (5) & & & & & -129 & (4) & & & & & -90& (1) & & & & & -90 & (1) & & & & & & & & \\
      & & & & BCQ$^\dagger$ & & & & -313 & (1) & & & & & -281 & (3) & & & & & -234 & (5) & & & & & -127 & (4) & & & & & -89 & (2) & & & & & & & & \\
      & & & & TD3+BC & & & & -325.9 & (1.9) & & & & & -289.4 & (1.6) & & & & & -230.1 & (2.1) & & & & & -172.9 & (1.9) & & & & & -112.5 & (2.1) & & & & & & & & \\
      & & & & CQL & & & & -291.8 & (3.1) & & & & & -327.1 & (0.2) & & & & & -326.3 & (2.1) & & & & & -322.6 & (1.7) & & & & & -271.9 & (1.6) & & & & & & & & \\
      & & & & MOPO & & & & \textbf{-123.8} & \textbf{(4.9)} & & & & & -110.3 & (4.3) & & & & & -139.7 & (8.4) & & & & & -130.7 & (5.2) & & & & & -119.4 & (4.9) & & & & & & & & \\
      & & & & MOReL & & & & -144.4 & (2.1) & & & & & -326.9 & (1.7) & & & & & -326.9 & (8.2) & & & & & -327.4 & (7.4) & & & & & -327.3 & (10.9) & & & & & & & & \\
      & & & & MOOSE$^\dagger$ & & & & -311 & (1) & & & & & -128& (1) & & & & & -110& (1) & & & & & -92.7 & (0.4) & & & & & -71.3 & (0.2) & & & & & & & &  \\
      & & & & WSBC$^\ddagger$ & & & & -134 & (2) & & & & & \textbf{-118}&\textbf{(1)} & & & & & \textbf{-103}& \textbf{(1)} & & & & & \textbf{-84.9} &\textbf{(0.2)} & & & & & \textbf{-70.0} & \textbf{(0.1)} & & & & & & & & \\

     \cline{4-36}
      &\parbox[t]{2mm}{\multirow{9}{*}{\rotatebox[origin=c]{90}{Mediocre}}} &
     & & BRAC-v$^\dagger$ & & & & -117 & (3) & & & & & -98.3 & (2.3) & & & & & -90.8 & (1) & & & & & -91.3 & (4.8) & & & & & -95.3 & (2.5) & & & & & -113 & (3) & & \\
     & & & & BEAR$^\dagger$ & & & & -111 & (1) & & & & & -115 & (7) & & & & & -109 & (4) & & & & & -111 & (6) & & & & & -104 & (3) & & & & & -65.1 & (0.3) & & \\
      & & & & BCQ$^\dagger$ & & & & -105 & (2) & & & & & -77.1 & (0.2) & & & & & -71.2& (0.3) & & & & & -78 & (1) & & & & & -125 & (4) & & & & & -68.6 & (0.3) & &\\
      & & & & TD3+BC & & & & -79.7 & (0.6) & & & & & -77.7 & (0.1) & & & & & -76.8& (0.1) & & & & & -74.3 & (0.1) & & & & & -70.7 & (0.1) & & & & & -65.4 & (0.1) & &\\
      & & & & CQL & & & & -88.9 & (1.3) & & & & & -80.5 & (0.3) & & & & & -80.8& (0.3) & & & & & -79.5 & (0.2) & & & & & -79.1 & (0.2) & & & & & -66.4 & (0.2) & &\\
      & & & & MOPO & & & & -102.4 & (3.6) & & & & & -119.1 & (5.2) & & & & & -81.4 & (2.3) & & & & & -86.1 & (2.5) & & & & & -92.5 & (4.5) & & & & & -105.1 & (7.4) & &\\
      & & & & MOReL & & & & -121.8 & (1.7) & & & & & -326.7 & (6.2) & & & & & -327.4 & (3.5) & & & & & -327.4 & (6.9) & & & & & -326.9 & (0.1) & & & & & -326.8 & (11.9) & &\\
     & & & & MOOSE$^\dagger$ & & & & -83.3& (0.3) & & & & & -76.6& (0.1) & & & & & -75.0 & (0.1) & & & & & \textbf{-71.1}& \textbf{(0.1)} & & & & & -69.7& (0.4) & & & & & -64.11& (0.02) & & \\
      & & & & WSBC$^\ddagger$ & & & & \textbf{-71.1}&\textbf{(0.2)} & & & & & \textbf{-68.5}&\textbf{(0.1)} & & & & & \textbf{-68.9} & \textbf{(0.1)} & & & & & -243& (1) & & & & & \textbf{-62.9}&\textbf{(0.2)} & & & & & \textbf{-63.76}&\textbf{(0.04)} & & \\
     
     \cline{4-36}
      &\parbox[t]{2mm}{\multirow{9}{*}{\rotatebox[origin=c]{90}{Optimized}}} &
     & & BRAC-v$^\dagger$ & & & & -127 & (5) & & & & & -78.4 & (0.9) & & & & & -165 & (6) & & & & & -76.9 & (2.1) & & & & & -98.7 & (3.1) & & & & & & & & \\
      & & & & BEAR$^\dagger$ & & & & -60.5 & (0.6) & & & & & -61.7 & (0.1) & & & & & -64.7 & (0.3) & & & & & -64.3 & (0.2) & & & & & -63.1 & (0.2) &  & & & & & & & \\
      & & & & BCQ$^\dagger$ & & & & -60.1 & (0.1) & & & & & -60.6 & (0.1) & & & & & -62.4 & (0.2) & & & & & -62.7 & (0.2) & & & & & -74.1 & (0.8) &  & & & & & & & \\
      & & & & TD3+BC & & & & -60.26 & (0.04) & & & & & -60.61 & (0.04) & & & & & -61.07 & (0.07) & & & & & -62.68 & (0.07) & & & & & -63.65 & (0.08) &  & & & & & & & \\
      & & & & CQL & & & & -60.88 & (0.09) & & & & & -60.55 & (0.11) & & & & & -60.65 & (0.09) & & & & & -60.61 & (0.09) & & & & & \textbf{-61.29} & (0.11) &  & & & & & & & \\
      & & & & MOPO & & & & -126.2 & (9.3) & & & & & -102.4 & (4.1) & & & & & -71.97 & (0.92) & & & & & -80.63 & (3.28) & & & & & -90.31 & (2.82) &  & & & & & & & \\
      & & & & MOReL & & & & -283.6 & (8.4) & & & & & -327.4 & (8.1) & & & & & -327.4 & (0.1) & & & & & -327.4 & (0.1) & & & & & -327.3 & (6.9) &  & & & & & & & \\
      & & & & MOOSE$^\dagger$ & & & & \textbf{-59.76}&\textbf{(0.02)} & & & & & -60.35& (0.02) & & & & & -60.77& (0.02) & & & & & -62.04& (0.02) & & & & & -62.73& (0.03) &  & & & & & & & \\
      & & & & WSBC$^\ddagger$ & & & & -60.18& (0.03) & & & & & \textbf{-58.2}& \textbf{(0.1)} & & & & & \textbf{-58.6}&\textbf{(0.1)} & & & & & \textbf{-59.39}&\textbf{(0.02)} & & & & & -61.70& (0.01) &  & & & & & & & \\
     \bottomrule
    \end{tabular}}
  \end{center}
\end{table*}

\section{Discussion / Conclusion}
When looking at the results presented in Table \ref{table:performance}, we can clearly see that the two purely model-based algorithms WSBC and MOOSE mostly outperform the other algorithms - the two exceptions being MOPO on the bad-0.0 dataset and CQL on the optimized-0.8 dataset.\\
In case of bad-0.0, it has to be noted though that the baseline is much worse than a random policy, i.e. almost everything an algorithm can do is better than strictly following the policy that generated this dataset, since also no exploration was added. It seems overall, that the hybrid methods were unable to accurately assess the model uncertainty correctly, which is not too surprising considering the 180-dimensional state and the complicated noise patterns of the industrial benchmark. The models would need to differentiate between aleatoric and epistemic uncertainty, which they are not intended for. Especially MOReL's USAD was thus not stopping the model rollouts soon enough - in our experiments we regularly experienced rollouts that were hundreds of steps long. Since no model can realistically be accurate enough for that long on the IB, the value function was trained on erroneous samples and the policy performs badly during evaluation. On all datasets except the $0\%$ exploration ones, MOReL's policys drive the system to one of the outermost edges of the steering state space - mostly without ever having seen any data there. MOPO's mistakes were not as grave, likely due to a combination of better models and much shorter rollouts. Its performances are however also rarely competetive. \\
Closely following WSBC and MOOSE are the model-free methods TD3+BC, BCQ, and CQL. They do not need to quantify uncertainty and are based on relatively simple regularization mechanisms, which appears to be an advantage compared to BEAR and BRAC. The latter methods both calculate divergences between the learned and behavioral policy - we hypothesize that this procedure is relatively unstable since the divergence measures have to be estimated from relatively few samples in order to be efficient (20 for BEAR, four for BRAC). In contrast TD3+BC, BCQ, and CQL find ways to incorporate samples from the behavior policy directly into policy regularization, without taking detours through action distributions (while a CQL variant additionally also penalizes the KL divergence between learned and behavior policy, we argue that the conservative Q-function plays the key role in this algorithm). We hypothesize further that this advantage is effectively transferred to the model-based methods MOOSE and WSBC. MOOSE uses a similar behavior policy representation as BCQ and also incorporates it much more directly than through divergence estimation into its policy training - simply by measuring reconstruction error of the new actions under the behavioral. WSBC has an even simpler mechanism in place, constraining the policy weights to not move outside an area closely around the behavioral policy's weights. We find thus that simplicity of the methods matters to performance.\\
WSBC and MOOSE seem to have a significant extra edge over TD3+BC, BCQ, and CQL, so the increased data efficiency of the model-based methods appears to be a real advantage in offline RL settings due to the natural data scarcity. The long rollouts are computationally more costly compared to the single step optimizations featured in model free algorithms' policy improvement steps. On the industrial benchmark, and we thus argue in many industrial settings, this does not matter too much though, because the effective planning horizon is not as extremely long as in some academic benchmarks, such as MuJoCo. In those benchmarks, transitions are perfectly deterministic, which is nearly never true in reality. Increased noise effectively limits the planning horizon that one can optimize for. MOOSE and WSBC can thus limit their optimization horizon to a realistic value and benefit from the advantages of increased data-efficiency in industrial offline RL.

\bibliography{ifacconf}             

\end{document}